\documentclass[10pt,twocolumn,letterpaper]{article}

\usepackage{cvpr}
\usepackage{times}
\usepackage{epsfig}
\usepackage{graphicx,subfigure}
\usepackage{amsmath}
\usepackage{amssymb}
\usepackage{amsthm}

\usepackage{multicol}
\usepackage{color}

\usepackage[table]{xcolor}% http://ctan.org/pkg/xcolor
\usepackage[pagebackref=true,breaklinks=true,letterpaper=true,colorlinks,bookmarks=false]{hyperref}

\cvprfinalcopy % *** Uncomment this line for the final submission

\def\calL{\mathcal{L}}

\def\s{\mathbf{s}}
\def\f{\mathbf{f}}

\def\a{\mathbf{a}}
\def\x{\mathbf{x}}
\def\h{\mathbf{h}}
\def\y{\mathbf{y}}
\def\z{\mbox{\boldmath$z$}}

\def\u{\mathbf{u}}

\def\LLambda{\mbox{\boldmath$\Lambda$}}

\newtheorem{problem}{Problem}
\newtheorem{theorem}{Theorem}

\definecolor{Gray}{gray}{0.85}
\definecolor{LightCyan}{rgb}{0.88,1,1}

\def\cvprPaperID{****} % *** Enter the CVPR Paper ID here

\ifcvprfinal\fi
\begin{document}

%%%%%%%%% TITLE
\title{Additive Adversarial Learning for Unbiased Authentication}

\author{Jian Liang$^{1*}$, Yuren Cao$^{1,2}$\thanks{Equal contribution from both authors.}, Chenbin Zhang$^{1,2}$, Shiyu Chang$^{3}$, Kun Bai$^{1}$, Zenglin Xu$^{2}$\\
$^1$Cloud and Smart Industries Group, Tencent, China\\
$^2$University of Electronic Science and Technology of China\\
$^3$MIT-IBM Watson AI Lab, IBM Research, USA\\
{\tt\small \{joshualiang,laurenyrcao,kunbai\}@tencent.com}\\
{\tt\small ChenbinZhang@std.uestc.edu.cn},
{\tt\small shiyu.chang@ibm.com},
{\tt\small zenglin@gmail.com}
}

\maketitle
%\thispagestyle{empty}

%%%%%%%%% ABSTRACT
\begin{abstract}
Authentication is a task aiming to confirm the truth between data instances and personal identities. Typical authentication applications include face recognition, person re-identification, authentication based on mobile devices and so on. The recently-emerging data-driven authentication process may encounter undesired biases, i.e., the models are often trained in one domain~(e.g., for people wearing spring outfits) while required to apply in other domains~(e.g., they change the clothes to summer outfits). To address this issue, we propose a novel two-stage method that disentangles the class/identity from domain-differences, and we consider multiple types of domain-difference. In the first stage, we learn disentangled representations by a one-versus-rest disentangle learning (OVRDL) mechanism. In the second stage, we improve the disentanglement by an additive adversarial learning (AAL) mechanism. Moreover, we discuss the necessity to avoid a learning dilemma due to disentangling causally related types of domain-difference. Comprehensive evaluation results demonstrate the effectiveness and superiority of the proposed method. 

\end{abstract}

%%%%%%%%% BODY TEXT
\section{Introduction}\label{sec:intro}

Authentication considers the problem of whether the data instances match personal identities. There is a variety of authentication applications including biometric authentication~\cite{bhattacharyya2009biometric,mir2011biometrics}~(e.g. face recognition~\cite{zhao2003face} and fingerprint verification~\cite{yager2004fingerprint}) and person re-identification~\cite{bedagkar2014survey,zheng2016person}. However, data-driven authentication process often suffers from undesired biases, i.e., domain-difference, which refers to the problem that a model is trained in one domain, but tested and verified in other domains. For example, in the field of person re-identification~\cite{bedagkar2014survey}, the prediction may be compromised due to the seasonal outfits changing or the angle variation between a camera and a pedestrian. 

\begin{table}\small
\begin{center}
\begin{tabular}{l|c|c|c}
\hline
\hline
 & Class Group 1 & Class Group 2 & Class Group 3\\
\hline\hline
Domain 1 & Train   & Test\cellcolor{Gray} & Test\cellcolor{Gray}\\
Domain 2 & Test \cellcolor{Gray}& Test\cellcolor{Gray} & Train \\
Domain 3 & Test \cellcolor{Gray}& Train  & Test\cellcolor{Gray}\\
\hline
\hline
\end{tabular}
\end{center}
\caption{An example of the assumptions of our problem.}\label{tab:proposed_problem}
\end{table}

Faced with the domain-difference problem between training and testing data, simply applying data-driven models may lead to undesired solutions that focus on the biases of domains, even if the training data is sufficient. To alleviate the aforementioned problem, this paper addresses the task of learning for unbiased authentication. For simplicity, we treat authentication as a recognition problem so that each identity corresponds to a class. We consider that there are multiple \emph{domains} and multiple types of \emph{domain-difference}, where a specific type of \emph{domain-difference} may include multiple \emph{domains}. For example, for person re-identification, \emph{season} and \emph{shooting angle} are two types of domain-difference, where \emph{season} includes four domains: \texttt{spring}, \texttt{summer}, \texttt{autumn}, and \texttt{winter}, and \emph{shooting angle} includes domains such as \texttt{front}, \texttt{back}, \texttt{side}, \emph{etc}.

\begin{figure*}[t]
\centering
\includegraphics[width=1\linewidth,height=6.5cm]{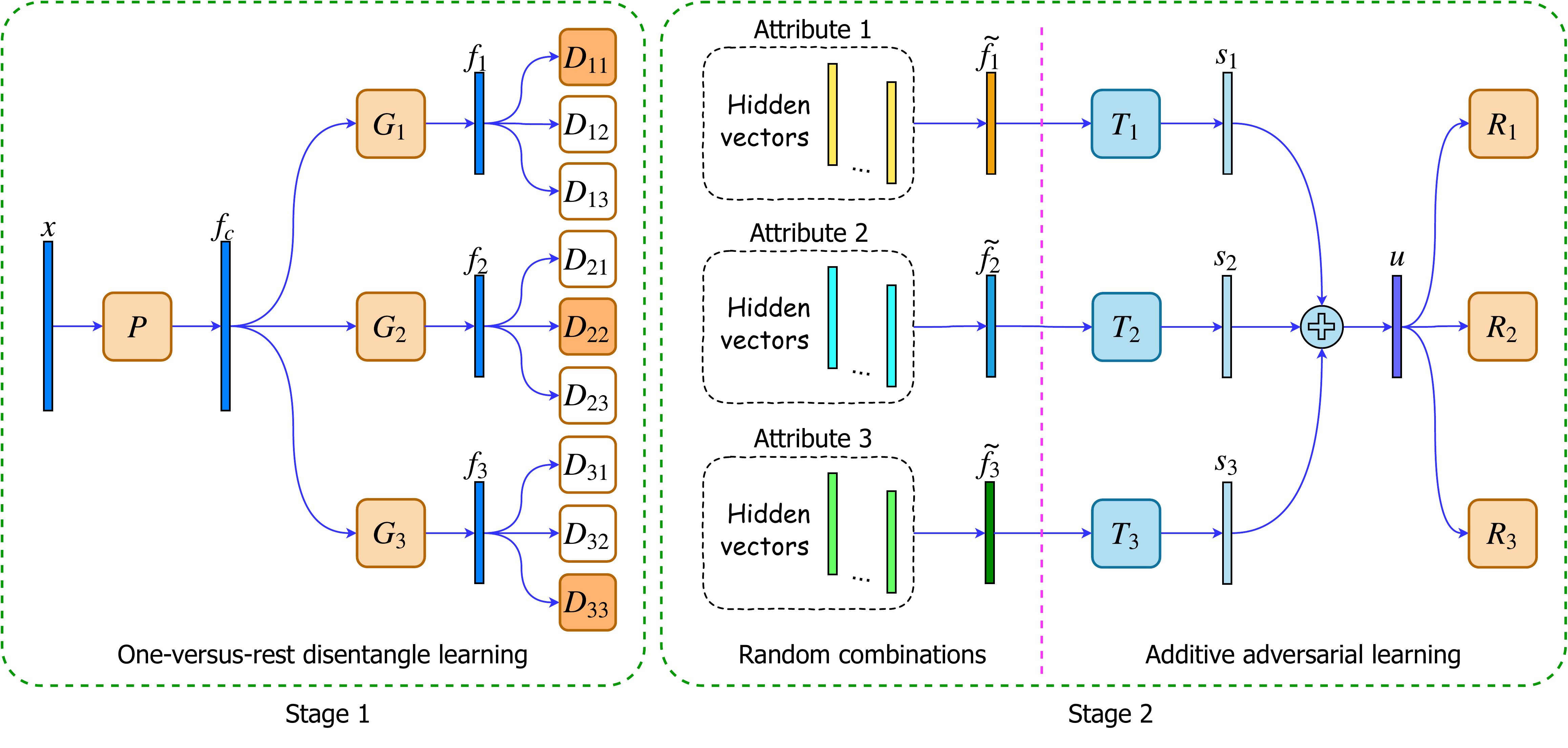}
\caption{The architecture of our framework. Intuitively, our framework is constructed in a multi-task learning flavor. The output of each task is regarded as an attribute to learn. The attribute-disentanglement pipeline of our work consists of two stages. Stage 1 consists of multiple branches of networks, and each branch learns by a one-versus-rest disentangling mechanism. Stage~2 aims to make further improvements, and the key ideas are illustrated in Fig.~\ref{fig:additive_learning}. Best viewed in color. }
\label{fig:model_architecture}
\end{figure*}

To better understand our problem, we present a toy example with only one type of domain difference in Table~\ref{tab:proposed_problem}. In the training phase, for each group of classes, we have their data on only one domain. In other words, different domains do not share classes. In the testing phase, we need to do a recognition on data which corresponds to unseen $\langle$class, domain$\rangle$ combinations. Mathematically, the problem we attempt to tackle is related to domain adaptation~\cite{patel2015visual,csurka2017domain,sun2015survey,wang2018deep}, but different from it, because domain adaptation allows source and target domains to share classes but provides no label on target domains. Domain adaptation has been extensively studied in the field of transfer learning~\cite{pan2010survey,weiss2016survey,wang2018deep,luo2018transfer}. Our problem can be transformed into a domain adaptation problem if the data of testing domains are allowed to train without class labels. Thus, we refer to our problem as a generalized cross-domain recognition (GCDR) problem. Similar problems have also been investigated in the field of fairness-oriented machine learning (FML) approaches~\cite{chouldechova2018frontiers} which concern biases against demographic groups, such as racial minorities or women. FML approaches in this setting usually apply transfer learning methods as solutions as well. In this paper, we also apply transfer learning methods to learn unbiased representations. Specifically, to focus on the main issue, we simply apply symmetric transfer learning methods (see the definition described by Weiss \etal~\cite{weiss2016survey}). 

In this paper, we propose a novel recognition method that learns disentangled representations to handle domain-difference to achieve an unbiased recognition. As shown in Table~\ref{tab:proposed_problem}, for a specific group of classes, the classes are different, but the domain is the same. Therefore, it is feasible to learn an unbiased model that can classify classes while neglecting the effects imposed by domain-differences. We also assume that although we have the labels of domains and domain-difference types, how the domain-differences affect the data is unknown. For a data instance, its class and domain values are treated as its attributes. Our method learns unbiased representations by disentangling these attributes. The framework of our method is illustrated in Fig.~\ref{fig:model_architecture}, which consists of two stages. In stage~1, we propose a one-versus-rest disentangle learning (OVRDL) mechanism to map each instance into multiple hidden spaces. In each hidden space, we disentangle one attribute from others. In stage 2, since limited combinations of attribute values are included in the training data, we conduct a data augmentation to randomly combine attribute labels and concatenate their associated hidden feature vectors as new data samples. An additive adversarial learning (AAL) mechanism based on random concatenations of hidden features is proposed to further improve the disentanglement of stage 1. Intuitively, biases are removed by minimizing negative side-effects. We extend the discussion on how to avoid a learning dilemma due to disentangle causally related attributes. The experimental results on benchmark and real-world data sets demonstrate the effectiveness and superiority of our method. We also conduct ablation experiments to show the contribution of each component of our proposed framework.

\section{Related Work}\label{sec:related_work}

% Apart from FML approaches, such a phenomenon of grouped classes is discussed by Bouchacourt \etal~\cite{bouchacourt2017multi} and Zhao~\cite{zhao2018network}. However, they do not provide learning methods eliminating domain-shifts. It is also discussed by Heinze-Deml and  Meinshausen~\cite{heinze2018grouping} that assumes classes with various domains are already included in training data. Yu \etal~\cite{yu2018multi} have also discussed the setting that classes are not necessarily shared by multiple source domains. They provide a method for the setting that all the $\langle$class, domain$\rangle$ combinations are included in training data set. 

To learn unbiased representations from unknown features of domain-difference, there are three thrusts of methods to leverage existing transfer learning methods, which are also the typical solutions for representation-learning based FML. In this section, we review them as well as some other related work, and differentiate them from our work.

\vspace{2mm}
\noindent\textbf{Eliminating the marginal-distribution differences \;}
The first family eliminates marginal-distribution differences between domains. This family of methods includes Transfer Component Analysis (TCA)~\cite{pan2011domain}, Deep Adaptation Network (DAN)~\cite{long2015learning}, Reversing Gradient (RevGrad)~\cite{ganin2014unsupervised}, Adversarial Discriminative Domain Adaptation (ADDA)~\cite{tzeng2017adversarial}, among others. FML methods proposed by Goel \etal~\cite{goel2018non} and Zhang \etal~\cite{zhang2018mitigating} also fall into this category. Many FML methods adopt RevGrad, such as those proposed by Wadsworth \etal~\cite{wadsworth2018achieving} and Beutel \etal~\cite{beutel2017data}.

% The first family adopts methods eliminating marginal distribution differences between different domains, such as Transfer Component Analysis (TCA)~\cite{pan2011domain}, Deep Adaptation Network (DAN)~\cite{long2015learning}, Reversing Gradient (RevGrad)~\cite{ganin2014unsupervised}, Adversarial Discriminative Domain Adaptation (ADDA)~\cite{tzeng2017adversarial}, among others. FML methods proposed by Goel \etal~\cite{goel2018non} and Zhang \etal~\cite{zhang2018mitigating} fall into this category. Many FML methods adopt RevGrad, such as those proposed by Wadsworth \etal~\cite{wadsworth2018achieving} and Beutel \etal~\cite{beutel2017data}.

\vspace{2mm}
\noindent\textbf{Generating data with unseen $\langle$class, domain$\rangle$ combinations \;}
The second family generates data samples associated with unseen $\langle$class, domain$\rangle$ combinations, such as ELEGANT~\cite{Xiao_2018_ECCV}, DNA-GAN~\cite{xiao2017dna}, Multi-Level Variational Autoencoder (ML-VAE)~\cite{bouchacourt2017multi}, CausalGAN~\cite{kocaoglu2017causalgan}, ResGAN~\cite{shen2017learning}, SaGAN~\cite{zhang2018generative}, among others. FML methods Fairness GAN~\cite{sattigeri2018fairness} and FairGAN~\cite{xu2018fairgan} also fall into this category. These methods generate synthetic data, then ordinary models can be trained on both real and the generated data.

% The second adopts generative methods which generate data samples associated with  unseen $\langle$class, domain$\rangle$ combinations, such as ELEGANT~\cite{Xiao_2018_ECCV}, DNA-GAN~\cite{xiao2017dna}, Multi-Level Variational Autoencoder (ML-VAE)~\cite{bouchacourt2017multi}, CausalGAN~\cite{kocaoglu2017causalgan}, ResGAN~\cite{shen2017learning}, SaGAN~\cite{zhang2018generative}, among others. FML methods Fairness GAN~\cite{sattigeri2018fairness} and FairGAN~\cite{xu2018fairgan} fall into this category. We can first adopt these methods to generate synthetic data, then to learn a direct feed forward network using both real and the generated data.

\vspace{2mm}
\noindent\textbf{Hybrid methods \;}
The third family performs both marginal-distribution-difference elimination and synthetic-data generation, such as Cross-Domain Representation Disentangler (CDRD)~\cite{liu2017detach}, Synthesized Examples for Generalized Zero-Shot Learning (SE-GZSL)~\cite{verma2018generalized}, Disentangled Synthesis for Domain Adaptation (DiDA)~\cite{cao2018dida}, Attribute-Based Synthetic Network (ABS-Net)~\cite{lu2018attribute}, among others. Madras \etal~\cite{madras2018learning} proposed such a FML framework.

% The third adopts methods performing both marginal-distribution-difference elimination and synthetic-data generation, such as Cross-Domain Representation Disentangler (CDRD)~\cite{liu2017detach}, Synthesized Examples for Generalized Zero-Shot Learning (SE-GZSL)~\cite{verma2018generalized}, Disentangled Synthesis for Domain Adaptation (DiDA)~\cite{cao2018dida}, Attribute-Based Synthetic Network (ABS-Net)~\cite{lu2018attribute}, among others. Madras \etal~\cite{madras2018learning} propose a FML method in this category.

\vspace{2mm}
\noindent\textbf{Other related work \;}
Such a phenomenon of grouped classes was also discussed by Bouchacourt \etal~\cite{bouchacourt2017multi} and Zhao~\cite{zhao2018network}. However, they did not provide learning methods to eliminate domain-differences. It was also discussed by Heinze-Deml and  Meinshausen~\cite{heinze2018grouping}. However, they assumed classes with various domains are already included in the training data. Yu \etal~\cite{yu2018multi} also discussed the setting that classes were not necessarily shared by multiple source domains. However, their method assumes that all the $\langle$class, domain$\rangle$ combinations are included in the training data set. 

\vspace{2mm}
\noindent\textbf{Differences between the existing works and our proposed method \;}
Despite the achievements, existing approaches either do not handle the GCDR problem or cannot avoid the learning dilemma due to disentangling correlated types of domain-difference. In addition, most of the generative methods generate samples in the original data space. However, if an appropriate model-structure is captured, generating data in the original data space is not necessary, and it may cause additional errors during both data generation and learning on the generated data. The aforementioned concerns are addressed by our proposed method. 

% However, these approaches do not provide solutions to handle multiple types of domain-shift with inter-type causalities. Therefore they may suffer either from negative transfer caused by forcing disentanglement, or from attribute entanglement caused by disentangling too few attributes. In addition, most of the generative methods aim to generate samples in the original data space. However, if an appropriate structure of the problem is captured, generating data in the original data space is not necessary. This will cause additional errors during both data generation and learning from the generated data. The aforementioned concerns are addressed by our proposed method.

\section{Methodology}\label{sec:method}

% \begin{figure}[t]
% \centering
% \includegraphics[width=1.0\linewidth]{co_mnist.jpg}
% \caption{Some examples of C-MNIST. Best viewed in color.}
% \label{fig:cmnist_dataset}
% \end{figure}

This section details our proposed network. We first define notations and problem settings.
Consider a data set $\mathcal{D} = \{(\x^i,y^i,\h^i)\}_{i=1}^n$ consisting of $n$ independent samples. For the $i$th sample, $\x^i\in\mathbb{R}^d$ is a feature vector with $d$ dimensions, $y^i\in\mathbb{Z}_+$ is a categorical class label of the recognition task, and $\h^i\in\mathbb{Z}_+^m$ is a vector consisting of $m$ categorical domain attributes.
For example, in the colored MNIST (C-MNIST) recognition (see the image examples in the Fig.~6 of Lu \etal~\cite{lu2018attribute}), $\x^i$ can be a colored image of digits with the size of $28\times 28$,  the class label denoted by $y^i$ is a value in $\{0,1,\ldots,9\}$, the background color (denoted by $h_1^i$) and foreground color (denoted by $h_2^i$) of the image are the two types of attributes. The different combinations of background colors and foreground colors can form multiple domains.
%Since the learning processes for the major recognition task and all the types of domain-difference are similar, we regard the class label and the domain labels of each sample as its attribute labels, and denote
For the convenience of the presentation, we denote $\a^i= (y^i,\h^i)\in\mathbb{Z}_+^{(m+1)}$ as the generalized attribute vector of the sample $i$. We denote $a_j^i$ as the $j$th element of $\a^i$, and  $a_j^i\in\{1,2,\ldots,k_j\}$, where $k_j$ is the  size of the set. Throughout the paper, we denote $[k]$ as the index set $\{1,2,\ldots,k\}$.

\begin{figure}[t]
\centering
\subfigure[Training]{\includegraphics[width=0.8\linewidth]{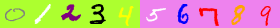}}
\subfigure[Testing]{\includegraphics[width=0.8\linewidth]{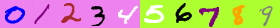}}
\caption{An experiment setting of C-MNIST with the background color as the domain-difference. Best viewed in color.}
\label{fig:cmnist_dataset_bg}
\end{figure}

In practice, samples of the data set $\mathcal{D}$ are usually incomplete. For the example shown in Fig.~\ref{fig:cmnist_dataset_bg}, one can observe images of digit $5$ with the red background, and digit $2$ with green background, while one wants to make predictions on images of $5$ with the green background. Formally, we define the GCDR problem as follows.

%The goal of our learning problem is to train a model over examples with partially observed combinations of attribute values, and then generalize this model to new examples with missing combinations of attribute values

\begin{problem}(Generalized Cross-Domain Recognition (GCDR))\label{def:proposed_problem}
Given a data set $\mathcal{D} = \{(\x^i,\a^i)\}_{i=1}^n$, %where $\x^i\in\mathbb{R}^d$, $\a^i\in\mathbb{Z}_+^{m+1}$ and for each element $\a_j^i\in\{1,2,\ldots,K_j\}$.
%For each $j$th type of domain-difference, let $k_j$ be the number of domains. 
let $\mathcal{D}_\Omega$ be the partially observed training set. 
The goal of our learning problem is to train a model over examples with partially observed combinations of attribute values, and then generalize this model to the testing set $\mathcal{D}_{\bar{\Omega}}$ with missing combinations of attribute values. 

Denote the sets of combinations of attribute values for the training and testing sets as $\mathcal{C}_\Omega=\{[a_1^i,\ldots,a_{(m+1)}^i]: i\in\Omega\}$ and $\mathcal{C}_{\bar{\Omega}}=\{[a_1^i,\ldots,a_{(m+1)}^i]: i\in\bar{\Omega}\}$, respectively. We have the constraint that the two sets have no intersection, \ie, $\mathcal{C}_\Omega\bigcap\mathcal{C}_{\bar{\Omega}} = \emptyset$. In addition, for the training set, for each $j$th type of domain-difference, denote the class group corresponding to its $r$th domain as $\mathcal{G}_{j}^r = \{y^i:h_j^i=r,r\in[k_j],i\in\Omega\}$. We have the constraint that different domains do not share classes, \ie, for each $j$th type of domain-difference, $\mathcal{G}_{j}^r\bigcap\mathcal{G}_{j}^{r'} = \emptyset$ for all $r,r'\in[k_j]$ and $r\neq r'$.
\end{problem}

%\begin{problem}\label{def:proposed_problem}
%Consider a data set $\mathcal{D} = \{(\x^i,y^i,\h^i)\}_{i=1}^n$, where $\x^i\in\mathbb{R}^d$, $y^i\in\mathbb{Z}_+$, and $\h^i\in\mathbb{Z}_+^m$.
%For each $j$th type of domain-difference, let $k_j$ be the number of domains. 
%Let $\Omega\subset[n]$ be the indices collection associated with observed samples which are for training. The samples associated with the missing indices are for testing.

%For each $j$th type of domain-difference, denote the class group corresponding to its $r$th domain as $\mathcal{G}_{j}^r = \{y^i:h_j^i=r,r\in[k_j],i\in\Omega\}$. We have the constraint that domains do not share classes, \ie, for each $j$th type of domain-difference, $\mathcal{G}_{j}^r\bigcap\mathcal{G}_{j}^{r'} = \emptyset$ for all $r,r'\in[k_j]$ and $r\neq r'$.
%\end{problem}

% training
%test

The structure of our framework is based on the ABS-Net~\cite{lu2018attribute}, and further novelly extends with these contributions: (1) a one-versus-rest disentangle learning mechanism, (2) an AAL mechanism to further improve disentangle performances, (3) an extended strategy to cease some disentangling processes to avoid a learning dilemma due to disentangling causally related attributes, which will be introduced in the following.

\subsection{One-Versus-Rest Disentangle Learning}\label{subsec:stage_1}

Our chief goal is to disentangle the class from all the types of domain-difference. Moreover, as an auxiliary, we also aim to disentangle each type of domain-difference from the class and other types of domain-difference. As mentioned previously, if we regard the class and all the types of domain-difference as attributes, we aim to disentangle each attribute from others. Therefore, we can develop a one-versus-rest strategy for each attribute to achieve two purposes: (1) to learn each attribute itself, and (2) to disentangle it from others.

Specifically, for each attribute $j\in[m+1]$, we learn the mapping from the raw-data space to a hidden space: $\x \rightarrow \f_j$ (here we omit the indices of sample order). In the hidden space, the two purposes above can be externalized as follows. (1) Samples associated with different categorical values of attribute $j$ can be well separated, \ie, $P(a_j\mid\f_j)=1$. (2) The distribution of samples is independent with that of any other attribute $j'$, \ie, $P(a_{j'}\mid\f_j)=P(a_{j'})$, which can be achieved by an adversarial learning process. 

%Such a modeling idea is common in transfer learning literature, such as those for RevGrad~\cite{ganin2014unsupervised} and ADDA~\cite{tzeng2017adversarial}.
 
As shown in Fig.~\ref{fig:model_architecture}, in stage 1, we construct a network to achieve the aforementioned purposes. For the feature vector $\x$ of a raw instance, it is transformed by an \textit{input-feature transformation network} $P$ into a hidden feature vector $\f_c$ which is further transformed by \textit{attribute-feature learning networks} $G_1,\ldots,G_{m+1}$ into \textit{attribute feature vectors} $\f_1,\ldots,\f_{m+1}$, respectively. For each attribute $j$, we expect the hidden space associated with the attribute feature vector $\f_j$ to achieve the two purposes above.

To achieve the aforementioned purposes, we develop a one-versus-rest disentangle learning (OVRDL) mechanism for each attribute. For each attribute $j$, we construct $(m+1)$ discriminative networks, $D_{j1},\ldots,D_{j(m+1)}$. Each discriminative network is trying to discriminate between different categorical values of the associated attribute. We expect that the ``diagonal network'' $D_{jj}$ learns directly and can correctly predict $a_j$, while the ``non-diagonal'' networks, $\{D_{jj'}\}_{j'\neq j}$, learn adversarially and cannot correctly predict $a_{j'}$. Following the adversarial learning mechanism proposed by Jolicoeur-Martineau~\cite{jolicoeur2018relativistic}, a brief version of the adversarial learning for the ``non-diagonal'' networks can be regarded as the following two alternative steps~\cite{jolicoeur2018relativistic}. 

\noindent \textbf{Step 1}: fix $G_j$, and for each $j'\neq j$, optimize $D_{jj'}$ to let the outputs approximate $\tilde{\a}_{j'}$ which is the one-hot-coded vector of the target $a_{j'}$;

\noindent \textbf{Step 2}: fix $D_{jj'}$s for all $j'\neq j$, and optimize $G_j$ to let the outputs approximate $(\mathbf{1} - \tilde{\a}_{j'})$.

%Given a matrix $\LLambda\in\mathbb{R}^{m\times m}$ consisting of causalities between attributes, then for each $j\in[m+1]$, we develop $m$ \textit{discriminative networks} $D_{j1},\ldots,D_{jm}$ for the $j$th attribute to handle those $m$ causalities with all the attributes.

% {\color{red}Although for each $j'\in[m+1]$, the targets for $D_{1j'},\ldots,D_{mj'}$ are the same, these discriminative networks do not share the same parameters, because (1) the parameter sharing will enforce $\f_1,\ldots,\f_m$ in the same space, which will increase the difficulties for them to learn independent vectors for independent attributes; (2) the causalities between attribute $j$ and $j'$ are different for different $j$s, thus the model cannot capture such causality structure due to parameter sharing.}

Finally, we establish the OVRDL mechanism in stage 1. For learning each attribute, we optimize by
\begin{equation}\label{eq:stage_1_task}\small
    \min_{P,\{G_j\},\{D_{jj}\}} \sum_{i\in{\Omega}}w_j\calL_{at}(D_{jj}(G_j(P(\x^i))),\tilde{\a}_j^i),
\end{equation}
where $\calL_{at}$ is the loss function for the attribute learning, $\tilde{\a}_j^i$ is the one-hot encoded vector of $a_j^i$, and $w_j$ is the weight for the $j$th attribute, $j\in[m+1]$.
For discriminating domains for each type of domain-difference, we optimize by
\begin{equation}\label{eq:stage_1_pos}\small
    \min_{\{D_{jj'}\}} \sum_{i\in{\Omega}}\sum_{j'\neq j}\tilde{w}_{jj'}\calL_{ad}(D_{jj'}(G_j(P(\x^i))),\tilde{\a}_{j'}^{i}),
\end{equation}
where $\calL_{ad}$ is the loss function for the adversarial learning, and $\tilde{w}_{jj'}$ is the weight for the $(j,j')$ pair, $j,j'\in[m+1]$.
To re-enforce attribute learning during the adversarial learning, we optimize by
\begin{equation}\label{eq:stage_1_neg_self}\small
    \min_{P} \sum_{i\in{\Omega}}\tilde{w}_{jj}\calL_{ad}(D_{jj}(G_j(P(\x^i))),\tilde{\a}_j^i).
\end{equation}
Finally, for eliminating all types of domain-difference, we optimize by
\begin{equation}\label{eq:stage_1_neg_other}\small
    \min_{P,\{G_j\}} \sum_{i\in{\Omega}}\sum_{j'\neq j}\tilde{w}_{jj'}\calL_{ad}(D_{jj'}(G_j(P(\x^i))),\tilde{\z}_{j'}^i),
\end{equation}
where $\tilde{\z}_{j'}^i = \mathbf{1} - \tilde{\y}_{j'}^i$.

The activation function chosen for the last layer of the discriminative networks is a softmax function. We choose the cross-entropy loss as $\calL_{at}$, and the mean square error as $\calL_{ad}$ (referring to LSGAN~\cite{mao2017least}). Eq.~\eqref{eq:stage_1_task},~\eqref{eq:stage_1_pos},~\eqref{eq:stage_1_neg_self} and~\eqref{eq:stage_1_neg_other} are alternatively optimized. For each mini-batch, Eq.~\eqref{eq:stage_1_task} and~\eqref{eq:stage_1_pos} run one step, while Eq.~\eqref{eq:stage_1_neg_self} and~\eqref{eq:stage_1_neg_other} run five steps.

For inference by only stage 1, we stack $P,G_1$ and $D_{11}$ to predict the class label $y_i$ for each sample $i$. Although disentangling the class from all the types of domain-difference can be accomplished only by the first branch of the network, \ie, the networks directly connected to $G_1$, we think that such a modeling strategy does not leverage sufficient supervised information to improve the representation ability of $P$. Later on in Section~\ref{sec:exp} we demonstrate that this results in a drastic decrease of accuracy by our ablation study. 

%  \begin{definition}[Demographic Parity]\label{th:demo_parity}
% A predictor $\hat{Y}$ satisfies \textit{demographic parity} if $\hat{Y}$ and a domain variable $Z$ are independent, which means $P(\hat{Y}\mid Z) = P(\hat{Y})$.
% \end{definition}

We show that our optimization scheme can improve \textit{Equality of Odds}, which is a fairness measure defined by Hardt \etal~\cite{hardt2016equality}. It means that a predictor $\hat{Y}$ and a domain variable $Z$ are conditionally independent given the true label $Y$, \ie, $P(\hat{Y}\mid Z,Y) = P(\hat{Y}\mid Y)$.

\begin{theorem}\label{th:independence}
For the GCDR problem and the defined model, our optimization scheme defined by Eq.~\eqref{eq:stage_1_task} $\sim$ Eq.~\eqref{eq:stage_1_neg_other} can improve the equality of odds defined by Hardt \etal~\cite{hardt2016equality}.
\end{theorem}
\begin{proof}
The proof is in the supplementary material.
\end{proof}

\subsection{Additive Adversarial Learning}\label{subsec:stage_2_additive}

To further improve the disentangle performance, we propose an additive adversarial learning (AAL) mechanism taking advantage of the attribute-combinations that are not seen in the training data. The attribute-combinations are generated by a data-augmentation procedure. We expect the AAL mechanism to have the following property: when the module ``sees'' the unseen $\langle$class, domain$\rangle$ combinations, biases are removed by minimizing negative side-effects.

First, we describe the data-augmentation procedure. For each $i$th generated data instance, the feature vector is a combination of $(m+1)$ feature vectors, $\tilde{\f}_{1}^i,\ldots,\tilde{\f}_{(m+1)}^i$, and the associated attribute vector is $[\tilde{a}_1^i,\ldots,\tilde{a}_{(m+1)}^i]$. For each attribute $j\in[m+1]$,  $\tilde{\f}_{j}^i = \f_j^{l}$ and $\tilde{a}_j^i = a_j^l$, where $\f_j^{l}$ and $a_j^l$ are the $l$th attribute feature vector and attribute value for attribute $j$, respectively, and $l\in\mathbb{Z}_+$ is a random index of training instances. For different attributes, the random indices can be different. For example, assuming $m = 1$, for two samples, $([\f_1^1,\f_2^1],[a_1^1,a_2^1])$ and $([\f_1^2,\f_2^2],[a_1^2,a_2^2])$, a generated data sample can be $([\f_1^1,\f_2^2],[a_1^1,a_2^2])$. The screening strategy of Lu \etal~\cite{lu2018attribute} is applied.

\begin{figure}[t]
\centering
\includegraphics[width=1\linewidth]{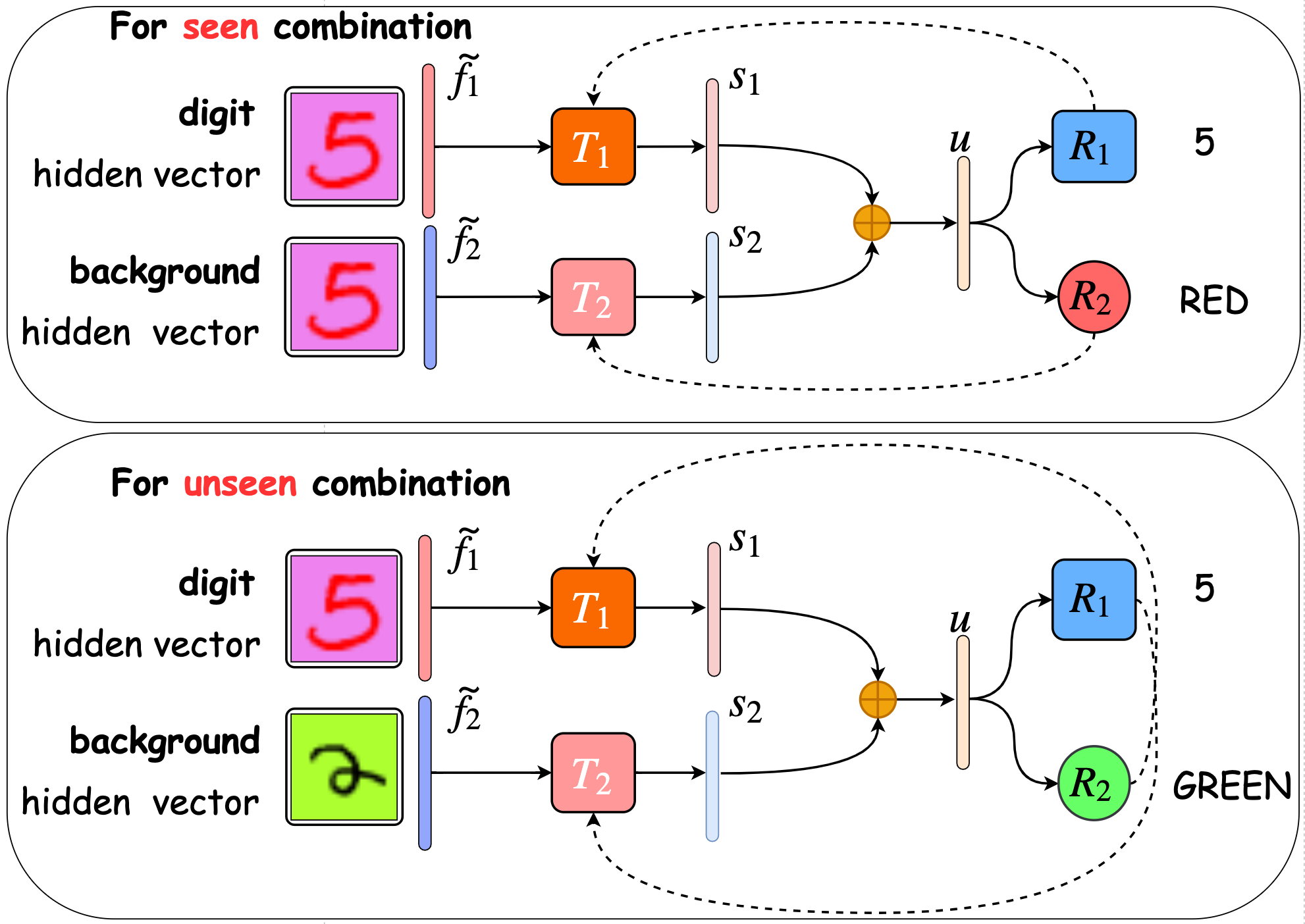}
\caption{Key ideas of our AAL mechanism. Dotted lines represent the directions of backpropagation. Best viewed in color.}
\label{fig:additive_learning}
\end{figure}

Next, we derive the AAL mechanism. The $n_r$ generated data samples are separated into two collections: $\Omega_{s}=\{i\in[n_r]$: the attribute-value combination $[\tilde{a}_1^i,\ldots,\tilde{a}_{(m+1)}^i]$ has been seen in the training data$\}$, and $\Omega_{u}=\{i\in[n_r]$: the attribute-value combination $[\tilde{a}_1^i,\ldots,\tilde{a}_{(m+1)}^i]$ has not been seen in the training data$\}$. Based on these two collections, we illustrate our key idea of AAL by Fig.~\ref{fig:additive_learning}. 
 Assume there are only two attributes: digit and background color, which are for the learning of two branches of the network, respectively. We assume that the disentanglement of stage~1 is already close to the optimum. Then for the seen attribute-value combinations, for each attribute $j$, we learn a transformation network $T_j$ to predict $\tilde{a}_j$ only. We assume that this learning process let the networks fit the data of seen combinations, \eg, a digit five with red background can be precisely recognized as ``5'' for digit and ``red'' for the background. Then for an unseen combination, a digit five and green background, we let the network to output ``5'' for digit and ``green'' for the background. Under the assumption above, if the output color is not ``green'', we believe the error is from the red information of the first branch. Therefore we back-propagate the loss from the second output to the first branch to eliminate the background information within. Finally, the bias in the first branch can be removed.

As shown in the second part of stage 2 in Fig.~\ref{fig:model_architecture}, for each generated data sample, the feature vectors, $\tilde{\f}_{1}^i,\ldots,\tilde{\f}_{(m+1)}^i$, are transformed into \textit{additive feature vectors} $\s_1,\ldots,\s_{(m+1)}$ by \textit{additive space transformation networks} $T_1,\ldots,T_{(m+1)}$, respectively. The additive feature vectors $\s_1,\ldots,\s_{(m+1)}$ are added as a \textit{summative feature vector} $\u$ which is sent to \textit{recognition networks} $R_1,\ldots,R_{(m+1)}$. For attribute-value combinations seen in the training data, for each attribute $j\in[m+1]$, the loss from $R_j$ is back propagated only to $T_j$, i.e., we optimize the following problem:
\begin{equation}\label{eq:stage_2_seen}\small
    \min_{R_j,T_j}  \sum_{i\in\Omega_s} w'_{j}\calL_{r}\biggl(R_{j}\biggl(\sum_{j'}T_{j'}(\tilde{\f}_{j'}^i)\biggr),\tilde{a}_{j}^i\biggr),
\end{equation}
where $\calL_{r}$ is the recognition loss function, and $w'_j$ is the weight for the attribute $j\in[m+1]$.
On the other hand, for attribute-value combinations unseen in the training data, for each attribute $j\in[m+1]$, the loss from $R_j$ is back propagated to $T_{\mathcal{S}_j}$, where $\mathcal{S}_j = \{j'\in[m+1]: j'\neq j\}$:
\begin{equation}\label{eq:stage_2_unseen}\small
    \min_{R_j,T_{\mathcal{S}_j}}  \sum_{i\in\Omega_u} w'_{j}\calL_{r}\biggl(R_{j}\biggl(\sum_{j'}T_{j'}(\tilde{\f}_{j'}^i)\biggr),\tilde{a}_{j}^i\biggr).
\end{equation}

%Here we still use the prior information of causalities to avoid disentangling correlated attributes based on Theorem~\ref{th:causality}.

The additive learning mechanism holds two good properties: (1) the discriminative information in each dimension can be expressed in an additive form which is decomposable, and (2) each dimension of $\s_j$ for all $j\in[m+1]$ has the same meaning, which allows us to incorporate sparse penalties to let each group of dimensions of the additive feature vectors correspond to a single attribute.

Same as in stage~1, we choose softmax activation functions and cross-entropy loss for the last layers. For inference, we stack $P,G_1,T_1$ and $R_1$ to predict the class label.

\subsection{Discussion on Causal Extension}

We further consider alleviating a dilemma of disentangling when some attributes are correlated. For the most extreme case, if two attributes are identical, it is not possible that we cannot recognize one attribute based on a feature vector but can recognize another. Therefore, intuitively, we should not disentangle correlated attributes. However, ``correlation'' is a broad, imprecise concept. If we do not disentangle for all the correlated attributes, we may encounter insufficient disentanglement. We consider a specific type of correlation: causal relationships. We theoretically demonstrate in Theorem~\ref{th:causality} that for any attribute $j$, if another attribute $j'$ causes it, then learning a feature vector $\f_j$ to recognize attribute $j$ while disentangling it from attribute $j'$ may hurt the recognition for attribute $j$. This is because if $\f_j$ is independent with attribute $j'$, since attribute $j'$ causes attribute $j$, the correlation between $\f_j$ and attribute $j$ is limited. Therefore, if the prior information of causal relationships between attributes is given, we should cease some disentangling processes to avoid the learning dilemma.

Specifically, for stage 1, we can use a prior matrix $\LLambda\in\{0,1\}^{(m+1)\times (m+1)}$ to handle causalities between attributes. For all $j,j'\in[m+1]$ and $j'\neq j$, we multiply the weight $\tilde{w}_{jj'}$ in Eq.~\eqref{eq:stage_1_pos} and Eq.~\eqref{eq:stage_1_neg_other} by $\LLambda_{jj'}$. Based on Theorem~\ref{th:causality}, we set $\LLambda_{jj'}=0$ if attribute $j'$ causes attribute $j$, and set $\LLambda_{jj'}=1$ otherwise. For stage 2, for the indices collection $\mathcal{S}_j$ in Eq.~\eqref{eq:stage_2_unseen}, we can delete the indices of attributes that are caused by attribute $j$, for each $j\in[m+1]$.

\begin{theorem}\label{th:causality}
For all attribute $j\in[m+1]$, for an arbitrary feature vector $\f_j$, denote the true label of $\f_j$ as $a_j\in\mathbb{Z}_+$. Then for attributes $j,j'\in[m+1]$ and $j\neq j'$, if attribute $j'$ causes attribute $j$, we cannot reach both the learning goals $P(a_j\mid\f_j)=1$ and $P(a_{j'}\mid\f_j)=P(a_{j'})$ perfectly. However, if attribute $j$ causes attribute $j'$, it is possible to reach both the learning goals perfectly. The proof can be found in the supplementary material.
\end{theorem}
\begin{proof}
The proof is in the supplementary material.
\end{proof}

\section{Experiments}\label{sec:exp}

In this section, we evaluate the proposed method. Both synthetic and real-world data sets are used for evaluations. Our implementation uses Keras with Tensorflow~\cite{abadi2016tensorflow} backends, which can be found at \href{https://github.com/langlrsw/AAL-unbiased-authentication}{https://github.com/langlrsw/AAL-unbiased-authentication}.

We consider a recognition task in the presence of several types of domain-difference. 
For each type of domain-difference, different domains do not share classes in the training set, and the training and testing sets do not share combinations of $\langle$class, domain$\rangle$. Comprehensive evaluations are conducted on three data sets: (1) the C-MNIST data set~\cite{lu2018attribute} with $10$ classes and $m=2$ types of domain-difference, (2) the re-organized CelebA data set~\cite{liu2015deep} with $211$ classes and $m=1$ type of domain-difference, and (3) our developed authentication data set based on mobile sensors with $29$ classes and $m=1$ type of domain-difference. For each data set, the re-organization will be described in detail in the corresponding section, and 10\% data of the testing set were randomly selected for validation. The evaluations follow the GCDR setting defined in Problem~\ref{def:proposed_problem}. 
 
\vspace{2mm}
\noindent\textbf{Methods for Comparison \;} As discussed in Section~\ref{sec:related_work}, there are three thrusts of methods to leverage existing transfer learning methods to handle domain-difference. For the first thrust that eliminates the marginal distribution differences, we chose RevGrad~\cite{ganin2014unsupervised}, which also serves as the solution of the FML methods of Beutel \etal~\cite{beutel2017data}. For the second thrust that generates data with unseen $\langle$class, domain$\rangle$ combinations, we chose ELEGANT~\cite{Xiao_2018_ECCV} which only uses domain labels and ML-VAE~\cite{bouchacourt2017multi} which only uses class labels. For the third thrust that uses hybrid solutions, we chose ABS-Net~\cite{lu2018attribute} which is the base method of ours without an adversarial mechanism, and CDRD~\cite{liu2017detach} and SE-GZSL~\cite{verma2018generalized}, which can be treated as advanced instantiated algorithms under the FML framework of Madras \etal~\cite{madras2018learning}. Finally, we compare the direct learning strategy that stacks $P,G_1$, and $D_{11}$ as the whole network.

\vspace{2mm}
\noindent\textbf{Evaluation Metrics \;}
We investigate prediction performances for both multi-label and multi-class types of recognition. Therefore, for the multi-label type, we use average AUC (aAUC) which is defined as the mean value of the area under the ROC curve for each class, the average false acceptance rate (aFAR), and the average false rejection rate (aFRR). Because the number of negative samples is far greater than that of positive samples for each class, we report aAUC and $(\mbox{aFAR}+\mbox{aFRR})/2$. For the multi-class type, we report top-1 accuracy (ACC@1).

% , average false acceptance rate (aFAR), average false rejection rate (aFRR), and average AUC (aAUC) which is defined as the mean value of the area under the ROC curve for each class. In this paper, for authentication, we mainly focus on  the ``$1:1$'' type of recognition (to answer ``Does this sample belong to the owner?''), not the ``$1:n$'' one (to answer ''Which owner does this sample belong to?''). Meanwhile, negative samples are far more than positive samples for each class/identity of the three data sets.
% Therefore, aAUC is considered as the main metric for comparison in our paper, and aACC results are omitted. $(\mbox{aFAR}+\mbox{aFRR})/2$ can also be used as a suitable metric for this setting. Nonetheless, our method is valid for the ``$1:n$'' type of recognition as well, hence ACC@1 is used for reference as well.

\vspace{2mm}
\noindent\textbf{Implementation Details \;}
For all experiments, $G_1,\ldots,G_m$ and $\{T_j\}$ are built by a single hidden layer with hyperbolic tangent as the activation function, respectively. Hidden units are flattened before being fed into {attribute-feature learning networks}. $\{D_{jj'}\}$ and  $\{R_j\}$ are built by generalized linear layers. A Convolutional Neural Network (CNN) with two convolutional layers is used as the input feature transformation network $P$ for image data sets. A fully-connected neural network with one hidden layer $P$ is used for vector based data sets. The weights $\{w_j\}$, $\{\tilde{w}_{jj'}\}$, and $\{w'_j\}$ are set as follows. We set $w_1 = w'_{1}= 1$ and $\tilde{w}_{j1} = 1$ for each $j\in[m+1]$. Other weights are equally set to $0.1$.

\subsection{Handwritten Digital Experiments}

We re-construct the C-MNIST data set originally built by Lu \etal~\cite{lu2018attribute} for performance evaluation. For original gray images of MNIST, $10$ different colors are added as background colors (b-colors) and other $10$ different colors are added as foreground colors (f-colors), which results in a new C-MNIST that consists of $70$k colored RGB digital images with resolution of $28\times 28$ ($60$k for training and $10$k for testing) from 1k possible combinations ($10$ digits $\times$ $10$ b-colors $\times$ $10$ f-colors). Examples from C-MNIST are shown in Fig.~6 of Lu \etal~\cite{lu2018attribute}.  

In this paper, we set the digit recognition as the primary learning task. The background and foreground colors can be treated as two types of domain-difference. It is evident that the background colors are independent with the digits. However, they have server influence on the prediction accuracy because they occupy most of the image area. Therefore, we use the background color as the domain-difference that groups the digits. As shown in Fig.~\ref{fig:cmnist_dataset_bg}, in the training data, digits $0\sim 4$ are associated with a green b-color, while $5\sim 9$ are associated with a pink b-color, other data are dropped. The data with $\langle 0\sim 4$, pink b-color$\rangle$ and $\langle 5\sim 9$, green b-color$\rangle$ combinations are for testing. The foreground attribute is also used to disentangle, but we allowed it share digits in the training data. We have $5970$ training instances and $1003$ testing instances in total.

Table~\ref{tab:result_basic_compare} summarizes the performance comparisons on C-MNIST. The results send a clear message that our method significantly outperforms the direct learning method, which shows the effectiveness of our method. Furthermore, our method outperforms other baseline methods significantly, except for the SE-GZSL method. We conjecture that C-MNIST is easy for SE-GZSL because the domain-difference is the background color which is simple and stable. However, in real applications, these properties barely hold, which on the succeeding two real-world data sets we will show that its performance drops.

We extended the experiments for other background colors and the foreground colors. Please find more details in the supplementary material.

\setlength{\tabcolsep}{2pt}
\begin{table}[t]\small
\begin{center}
\begin{tabular}{l|ccc}
\hline
\hline
 Methods  & aAUC  & $(\mbox{aFAR}+\mbox{aFRR})/2$ & ACC@1 \\\hline\hline
 Direct & 78.35     & 26.70& 20.96\\
 RevGrad~\cite{ganin2014unsupervised,beutel2017data}     & 80.71  &24.45 & 21.68 \\
CDRD~\cite{liu2017detach,madras2018learning} & 84.83  & 35.79  & 33.49 \\
SE-GZSL~\cite{verma2018generalized,madras2018learning} &  \textbf{99.79}    & \textbf{2.72}& \textbf{94.83} \\
 ELEGANT~\cite{Xiao_2018_ECCV}    &  79.94    & 24.61&10.68 \\
 ML-VAE~\cite{bouchacourt2017multi}  &77.26    & 28.06 & 18.73\\
 ABS-Net~\cite{lu2018attribute} & 77.69    &27.41& 15.92 \\\hline
 Ours &  98.42  & 6.14 & 84.27 \\\hline\hline
\end{tabular}
\end{center}
\caption{Performances (\%) comparison on the C-MNIST data set. ``Direct'' means stacking $P,G_1$, and $D_{11}$ as the whole network.}\label{tab:result_basic_compare}
\end{table}

\subsection{Face Recognition}

We use aligned, and cropped version of the CelebA data set~\cite{liu2015deep} and scale all images to $64\times 64$. We chose the \textit{Eyeglasses} attribute as the domain-difference. We select individuals who have at least $20$ images and $\#(Eyeglasses=0)/\#(Eyeglasses=1)\in[3/7,7/3] $, resulting in $211$ individuals.
Half of the individuals wear glasses during training and without glasses during testing. The other half wear no glasses during training and wear glasses during testing. 
Table~\ref{tab:result_celebA_compare} shows the comparisons conducted on CelebA. Our method significantly outperforms other methods in aAUC and $(\mbox{aFAR}+\mbox{aFRR})/2$---the multi-label type of metrics. The SE-GZSL method underperforms, which suggests its insufficient inconsistency. The result demonstrates that complex and variable domain-difference types on real-world data sets are difficult for SE-GZSL to learn. The CDRD method underperforms in aAUC and $(\mbox{aFAR}+\mbox{aFRR})/2$, but outperforms in ACC@1, because the positive samples of the majority of individuals have lower prediction scores, but the positive samples of more individuals have high prediction scores, which shows less satisfactory authentication performance for the majority of individuals.
  
\setlength{\tabcolsep}{2pt}
\begin{table}[t]\small
\begin{center}
\begin{tabular}{l|ccc}
\hline\hline
 Methods  & aAUC  & $(\mbox{aFAR}+\mbox{aFRR})/2$& ACC@1  \\\hline\hline
 Direct & 78.54    & 42.10& 11.07 \\
 RevGrad~\cite{ganin2014unsupervised,beutel2017data}     & 80.12   &  31.18& 10.96\\
CDRD~\cite{liu2017detach,madras2018learning} & 80.20    &   39.90& \textbf{16.47}\\
SE-GZSL~\cite{verma2018generalized,madras2018learning}  & 84.96   &26.62 & 12.76
\\
 ELEGANT~\cite{Xiao_2018_ECCV}    &  75.88  & 32.02&  10.05\\
 ML-VAE~\cite{bouchacourt2017multi}     & 75.29    &36.07& 7.97 \\
 ABS-Net~\cite{lu2018attribute} & 75.80    &34.90& 8.09\\\hline
 Ours &  \textbf{87.07}   & \textbf{22.19}&  {14.99}\\\hline\hline
\end{tabular}
\end{center}
\caption{Performances (\%) comparison on the CelebA data set.}\label{tab:result_celebA_compare}
\end{table} 

\subsection{Authentication on Mobile Devices}\label{subsubsec:PIE}

We also build a data set containing sensor information of smart-phones from 29 subjects, which records two-second time-series data from multiple sensors, such as accelerometer, gyroscope, gravimeter, \etc. Statistical features from both time and spectrum domains are extracted with 191 dimensions for all the 5144 data instances. The OS types (IOS/Android) are considered to be the domain-difference. We select 12 subjects that have used both types of the phone system and then construct a biased learning task as shown in Table~\ref{tab:mobile_problem}. The ELEGANT method is not suitable for non-image data and therefore is removed. The results are reported in Table~\ref{tab:result_mobile_compare}, in which our method significantly outperforms other methods in aAUC and $(\mbox{aFAR}+\mbox{aFRR})/2$. The SE-GZSL method also underperforms for this difficult type of domain-difference as well. The CDRD method still underperforms in the multi-label type of metrics, especially in $(\mbox{aFAR}+\mbox{aFRR})/2$.

\begin{table}[t]\small
\begin{center}
\begin{tabular}{c|c|c|c|c}
\hline\hline
 & No. 1-6 & No. 7-12 & No. 13-15&No. 16-29\\
\hline\hline
IOS & Train   & Test & $ \times$ & Train\\\hline
Android & Test  & Train  & Train&  $ \times$\\
\hline\hline
\end{tabular}
\end{center}
\caption{The authentication problem on mobile devices. The numbers in the first row indicate groups of subjects. ``$ \times$'' means there are no data for this condition.}\label{tab:mobile_problem}
\end{table}

\setlength{\tabcolsep}{2pt}
\begin{table}[t]\small
\begin{center}
\begin{tabular}{l|ccc}
\hline\hline
 Methods  & aAUC  &   $(\mbox{aFAR}+\mbox{aFRR})/2$& ACC@1  \\\hline\hline
 Direct & 76.90    & 28.21& 3.56  \\
 RevGrad~\cite{ganin2014unsupervised,beutel2017data}     & 75.88   &32.38& 0.38 \\
CDRD~\cite{liu2017detach,madras2018learning} & 89.17    &20.26& 46.05 \\
SE-GZSL~\cite{verma2018generalized,madras2018learning}  & 78.83   &26.12& 20.54  \\
 ML-VAE~\cite{bouchacourt2017multi}     &77.16    & 27.18& 4.68  \\
 ABS-Net~\cite{lu2018attribute} &76.58     & 28.09& 5.13\\\hline
 Ours &  \textbf{93.40}    & \textbf{13.59}&  \textbf{46.37}\\\hline\hline
\end{tabular}
\end{center}
\caption{Performances (\%) comparison on the Mobile data set.}\label{tab:result_mobile_compare}
\end{table}
   
\subsection{Ablative Study}\label{subsec:ablation}
 
We conduct a series of ablation experiments on the C-MNIST data set in the aforementioned setting to demonstrate how the OVRDL (stage 1) and AAL (stage 2) mechanisms contribute to the performance. Specifically, we compare the performance of the following four model variants.

\textit{Single-Branch}. Only the first branch of networks in stage 1 is left. Stage 2 is therefore removed because it works with a multi-branch stage 1. 

\textit{Shared-$D$s}. $D_{1j}  =\ldots=D_{mj}$, for all $j\in[m+1]$.

\textit{No-Adv-Stage-1}. The networks $\{D_{jj'}\}_{j\neq j'}$ in stage 1 are removed. 

\textit{No-Adv-At-All}. Based on No-Adv-Stage-1, in stage 2, losses are back-propagated to all the networks as normal. It is the ABS-Net~\cite{lu2018attribute} method.
 
\setlength{\tabcolsep}{2pt}
\begin{table}[t]\small
\begin{center}
\begin{tabular}{l|ccc}
\hline\hline
 Methods  & aAUC  & $(\mbox{aFAR}+\mbox{aFRR})/2$ & ACC@1  \\\hline\hline
 Stage 1+2 &   \textbf{98.42 }   &\textbf{6.14}& \textbf{84.27} \\\hline
 Stage 1 &  95.91    & 10.20& 70.56 \\
 Single-Branch & 78.87    &26.70&  24.47 \\
 Shared-$D$s & 87.27   &18.72& 46.75 \\
 No-Adv-Stage-1 & 92.47    & 14.81&  46.97\\
 No-Adv-At-All & 77.69    &27.41& 15.92\\
 Direct & 78.35     & 26.70& 20.96\\\hline\hline
\end{tabular}
\end{center}
\caption{Results of the ablation study.}\label{tab:result_ablation}
\end{table}

The results are presented in Table~\ref{tab:result_ablation}. 
It is notable that Single-Branch's performances drastically decrease comparing with other methods containing adversarial learning. The performances of Single-Branch are similar to those of the related approaches listed in Table~\ref{tab:result_basic_compare}. Such phenomenon suggests that building a single-branch model to handle domain-differences is not sufficient and that it is worthwhile to build a multi-branch model to learn all the attributes to improve the representation ability of $P$. Such a multi-branch structure is the main difference between our method and the related approaches, which we believe is of the main structural contributions of our framework. Besides, Shared-$D$s' performances also decrease considerably. It is worth mentioning that, for Shared-$D$s, the performances of both stages are nearly the same. These phenomena demonstrate that restricting the attribute vectors in the same space harms their learning of independent representations. Comparing No-Adv-Stage-1 with Shared-$D$s, no adversarial learning in stage 1 is better than ``shared'' adversarial learning, which also demonstrates the importance of independent representations. These phenomena demonstrate the high effectiveness of our proposed OVRDL mechanism.
 
On the other hand, compared stage 1+2 with stage 2, the improvement gained from stage 2 is significant. For No-Adv-Stage-1, it is worth mentioning that it can only achieve aAUC of 75\% without stage 2. Moreover, No-Adv-At-All significantly underperforms compared with No-Adv-Stage-1 by only changing the back-propagation mechanism of stage 2. These phenomena demonstrate the high effectiveness of our proposed AAL mechanism in stage 2.
 
\begin{figure}
\begin{center}
\includegraphics[width=1.0\linewidth,height=4cm]{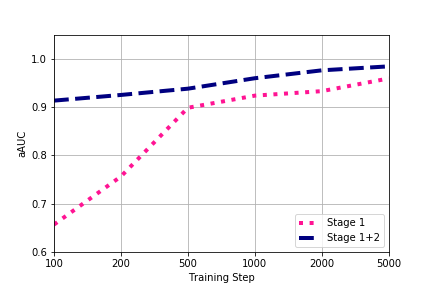}
\end{center}
   \caption{Improvements of the AAL mechanism in stage 2 during different training phases of stage 1.}
\label{fig:behavior_stage_2}
\end{figure}

We further investigate the effectiveness of the AAL mechanism during different training phases of stage 1.  Fig.~\ref{fig:behavior_stage_2} shows that the AAL mechanism can contribute 25\% absolute performances at the beginning of the training. At the middle and later phases of training, the improvements are limited, because AAL aims to eliminate biased factors in the features further, but such factors are nearly cleansed to the optimum by stage~1.

% \subsection{Stability for Different Domain-Differences}
% To investigate the stability of our proposed method, we conduct experiments with various domain-differences on the C-MNIST data set. We consider both the background color and the foreground color as the domain-differences that group digits. For both background and foreground colors, we randomly select $10$ combinations of different two colors as two domains. The average performances reported in Table~\ref{tab:result_avg_diff_domain} show stable performances for both our OVRDL (stage 1) and our AAL (stage 2) mechanisms.

% \setlength{\tabcolsep}{2pt}
% \begin{table}[t]\small
% \begin{center}
% \begin{tabular}{l|ccc}
% \hline\hline
%  Methods  & aAUC &  $(\mbox{aFAR}+\mbox{aFRR})/2$& ACC@1    \\\hline\hline
%  F-Colors, Direct & 66.61    &41.56  & 10.06\\
%  F-Colors, Stage 1 & 87.59     & 18.95& 42.33\\
%  F-Colors, Stage 1+2 & \textbf{92.93}    & \textbf{13.61}& \textbf{58.32}  \\\hline\hline
%  B-Colors, Direct &79.63    & 30.46 & 25.47  \\
%  B-Colors, Stage 1 &94.38     & 11.75& 62.78  \\
%  B-Colors, Stage 1+2 &\textbf{96.32}      &  \textbf{9.56} & \textbf{68.83}\\\hline\hline
% \end{tabular}
% \end{center}
% \caption{Average performances (\%) of different domain-differences on the C-MNIST data set. F-color and B-color stand for foreground and background colors, respectively. }\label{tab:result_avg_diff_domain}
% \end{table}
 
\section{Conclusion}\label{sec:conclusion}
In this paper, we investigate data biases and a generalized cross-domain recognition problem in the field of authentication where domains do not share classes. We recognize the class for unseen $\langle$class, domain$\rangle$ combinations of data. We propose a two-stage disentangle learning method to tackle the problem. The stage 1 builds a one-versus-rest disentangle learning mechanism to disentangle the class and each type of domain-difference. The stage 2 conducts a data augmentation and uses a proposed additive adversarial learning to improve the disentanglement of stage 1 further. We also discuss how to avoid the dilemma due to disentangling causally related types of domain-difference. The experiments demonstrate that our method significantly outperforms existing state-of-the-art methods. We also conduct an ablation study to demonstrate the effectiveness of the critical components of our method. Some interesting future directions of research include developing transfer learning algorithms flexible to train to increase the number of types of domain-difference.

\section*{Acknowledgment}
 
We would like to thank the TuringShield team of Tencent for supporting our research, Lu \etal~\cite{lu2018attribute} for sharing their codes of ABS-Net and C-MNIST data set, and Dr. Bing Bai from the Cloud and Smart Industries Group at Tencent for his insightful suggestions.

\newpage

{\small
\bibliographystyle{ieee}
\bibliography{main}
}

\end{document}